\documentclass[journal]{IEEEtran}

\usepackage{epsfig}
\usepackage{graphicx}
\usepackage{amsmath}
\usepackage{amssymb}

\usepackage{xcolor}
\usepackage{soul}
\usepackage[utf8]{inputenc}
\usepackage[pagebackref=true,breaklinks=true,letterpaper=true,colorlinks,bookmarks=false,citecolor=cyan]{hyperref}

\usepackage[figuresright]{rotating}
\usepackage{textcomp,subfigure,algorithm,algorithmic,multirow,upgreek,tabularx}
\usepackage{graphics,gensymb}
\usepackage{bm,cases,threeparttable}

\usepackage{setspace}

\graphicspath{{Figures/}}
\usepackage{booktabs}

\usepackage{amsfonts}
 
\usepackage{caption}
\usepackage{graphicx} %
\usepackage{sidecap}

\usepackage{enumitem}
\setitemize{noitemsep,topsep=0pt,parsep=0pt,partopsep=0pt}

\ifCLASSINFOpdf
\else
\fi

\begin{document}
%
\title{Total Generate: Cycle in Cycle Generative Adversarial Networks for Generating Human Faces, Hands, Bodies, and Natural Scenes}


\author{\IEEEauthorblockN{Hao Tang and
		Nicu Sebe
}
\thanks{
Hao Tang and Nicu Sebe are with the Department of Information Engineering and Computer Science (DISI), University of Trento, Trento 38123, Italy. E-mail: hao.tang@unitn.it, sebe@disi.unitn.it.
}
}

\markboth{IEEE Transactions on Multimedia}%
{Shell \MakeLowercase{\textit{et al.}}: Bare Demo of IEEEtran.cls for IEEE Transactions on Magnetics Journals}
%

\IEEEtitleabstractindextext{%

\begin{abstract}
We propose a novel and unified Cycle in Cycle Generative Adversarial Network (C2GAN) for generating human faces, hands, bodies, and natural scenes.
Our proposed C2GAN is a cross-modal model exploring the joint exploitation of the input image data and guidance data in an interactive manner. 
C2GAN contains two different generators, i.e., an image-generation generator and a guidance-generation generator. Both generators are mutually connected and trained in an end-to-end fashion and explicitly form three cycled subnets, i.e., one image generation cycle and two guidance generation cycles.
Each cycle aims at reconstructing the input domain and simultaneously produces a useful output involved in the generation of another cycle. 
In this way, the cycles constrain each other implicitly providing complementary information from both image and guidance modalities and bringing an extra supervision gradient across the cycles, facilitating a more robust optimization of the whole model. 
Extensive results on four guided image-to-image translation subtasks demonstrate that the proposed C2GAN is effective in generating more realistic images compared with state-of-the-art models.
The code is available at \url{https://github.com/Ha0Tang/C2GAN}.
\end{abstract}

}

\maketitle

\IEEEdisplaynontitleabstractindextext

%
\IEEEpeerreviewmaketitle

\section{Introduction}

In this work, we focus on how to generate a target image given an input image. This has many application scenarios such as human-computer interaction, entertainment, virtual reality, and data augmentation. However, this task is challenging since it needs a high-level semantic understanding of the image mapping between the input and the output domains. Recently, Generative Adversarial Networks (GANs)~\cite{goodfellow2014generative} have shown the potential to solve this challenging task. GANs have produced promising results in many tasks such as image generation~\cite{tang2020unified}, image inpainting~\cite{zhang2020dual}, and cross-modal translation \cite{duan2019cascade}.

Recent works have developed powerful image-to-image translation systems, e.g., Pix2pix~\cite{isola2016image} and GauGAN~\cite{park2019semantic} in supervised settings, and CycleGAN~\cite{zhu2017unpaired} and DualGAN~\cite{yi2017dualgan} in unsupervised settings.
However, these methods are tailored to merely two domains at a time and scaling them to more requires a quadratic number of models to be trained. 
For instance, with $m$ different image domains, CycleGAN and Pix2pix need to train $m(m{-}1)/2$ and $m(m{-}1)$ models, respectively.
To overcome this, Choi et al. propose StarGAN~\cite{choi2017stargan}, in which a single generator/discriminator performs image-to-image translation for multiple domains. 
However, StarGAN is not effective in handling some specific image-to-image translation tasks such as human pose generation~\cite{ma2017pose,siarohin2017deformable}, hand gesture generation~\cite{tang2018gesturegan}, and cross-view image translation \cite{regmi2018cross}, in which image generation could involve infinite image domains since human body, hand gesture, and natural scene in the wild can have arbitrary poses, sizes, appearances, locations, and viewpoints.

\begin{figure*}[!t] \small
	\centering
	\includegraphics[width=.9\linewidth]{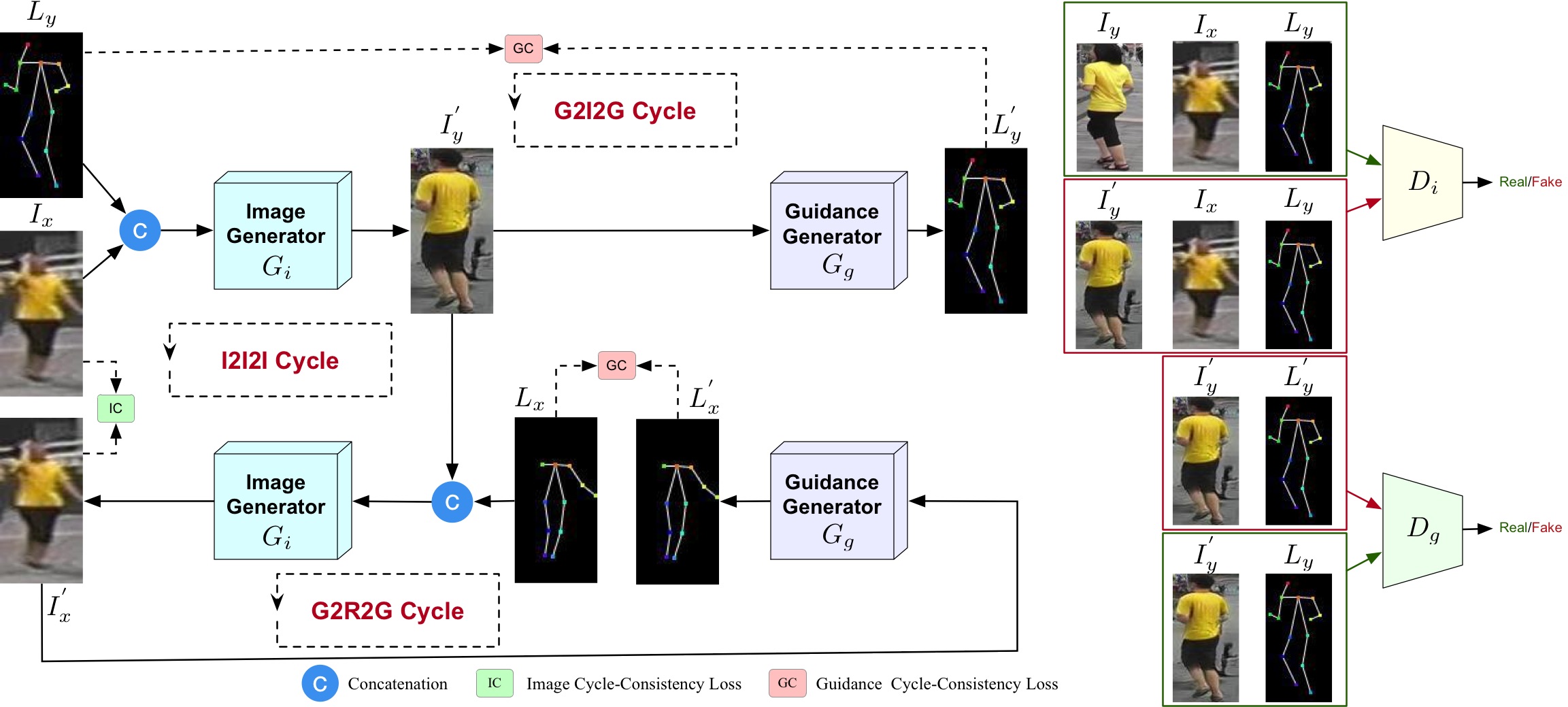}
	\caption{Overview of the proposed C2GAN, which consists of two types of generators, i.e., image generator $G_i$ and guidance generator $G_g$.  Parameter-sharing strategies can be used in between the image or the guidance generators to reduce the model capacity. 
	During the training stage, two generators $G_i$ and $G_g$ are explicitly connected and trained by three cycles, i.e., the image cycle I2I2I: $[I_x, L_y] \stackrel{G_i}\rightarrow [I_y^{'}, L_x] \stackrel{G_i}\rightarrow I_x^{'}$ and two guidance cycles G2I2G: $[I_x, L_y] \stackrel{G_i}\rightarrow I_y^{'} \stackrel{G_g}\rightarrow L_y^{'}$, G2R2G: $[I_y^{'}, L_x] \stackrel{G_i}\rightarrow I_x^{'} \stackrel{G_g}\rightarrow L_x^{'}$. The right side of the figure shows the cross-modal discriminators (i.e., $D_i$ and $D_g$) for a better network optimization.
	} 
	\label{fig:excyclegan}
	\vspace{-0.4cm}
\end{figure*}

To address these limitations, many methods are proposed to generate images based on extra semantic guidance, such as object keypoints \cite{song2017geometry,ma2017pose}, human skeletons \cite{siarohin2017deformable,tang2018gesturegan}, or segmentation maps \cite{regmi2019cross,tang2019multi,park2019semantic,tang2019local,tang2020edge}. 
For instance, Song et al.~\cite{song2017geometry} propose a G2GAN framework for facial expression synthesis based on facial landmarks.
Siarohin et al.~\cite{siarohin2017deformable} introduce a PoseGAN model for pose-based human image generation conditioned on human body skeletons.
Regmi and Borji \cite{regmi2018cross} propose both X-Fork and X-Seq for cross-view image translation conditioned on segmentation maps. 
However, the current state-of-the-art guided image-to-image translation methods such as PG2~\cite{ma2017pose}, PoseGAN~\cite{siarohin2017deformable}, X-Fork \cite{regmi2018cross}, and X-Seq \cite{regmi2018cross} have two main issues: 1) they directly transfer a source image and the target guidance to the target domain (i.e., $[I_x, L_y] \stackrel{G_i} \rightarrow I_y^{'}$ in Fig.~\ref{fig:excyclegan}), without considering the mutual translation between each other, while the translation across different image and guidance modalities in a unified framework would bring rich cross-modal information;
2) they simply employ the guidance data as input to guide the generation process, without involving the generated guidance as supervisory signals to further improve the network optimization. Both issues lead to unsatisfactory results.

To address both, we propose a novel and unified Cycle In Cycle Generative Adversarial Network (C2GAN), in which three cycled sub-nets are explicitly formed to learn both image and guidance modalities in a joint model. 
The framework of the proposed C2GAN is shown in Fig. \ref{fig:excyclegan}. 
Specifically, to address the first limitation, C2GAN contains an image cycle, i.e., I2I2I ($[I_x, L_y] \stackrel{G_i}\rightarrow [I_y^{'}, L_x] \stackrel{G_i} \rightarrow I_x^{'}$), which aims at reconstructing the input and further refines the generated images $I_y^{'}$.
To address the second limitation, the guidance information (such as the human body skeleton) in C2GAN is not only utilized as input but also acts as output, meaning that the guidance is also a generative objective. The input and output of the guidance are connected by two novel guidance cycles, i.e., G2I2G ($[I_x, L_y] \stackrel{G_i}\rightarrow I_y^{'} \stackrel{G_g}\rightarrow L_y^{'}$) and G2R2G ($[I_y^{'}, L_x] \stackrel{G_i}\rightarrow I_x^{'} \stackrel{G_g}\rightarrow L_x^{'}$), where $G_i$ and $G_g$ denote an image and a guidance generator, respectively. In this way, guidance cycles can provide weak supervision to the generated images $I_y^{'}$. The intuition behind the guidance cycles is that if the generated guidance is very close to the real guidance, then the corresponding images should be similar (see Fig.~\ref{fig:guidance}). 
In other words, a better guidance generation will boost the performance of image generation, and conversely the improved image generation will further facilitate the guidance generation. 
The proposed three cycles inherently constraint each other in an end-to-end training fashion. Moreover, for a better optimization of the proposed three cycles, we further propose two novel cycle losses, i.e., Image Cycle-consistency loss (IC) and Guidance Cycle-consistency loss (GC). With both cycle losses, each cycle can benefit from each other in a joint learning way. We also propose two cross-modal discriminators corresponding to the generators. 

Our contributions can be summarized as follows:
\begin{itemize}[leftmargin=*]
	\item We propose C2GAN, a novel and unified cross-modal generative adversarial network for guided image-to-image translation tasks, which organizes the guidance and image data in an interactive manner, instead of using as input only the guidance information.
	\item The proposed cycle in cycle network structure is a new design which explores the effective use of cross-modal information for guided image-to-image translation tasks. The designed cycled subnetworks connect different modalities and implicitly constraint each other, leading to extra supervision signals for better image generation. We also investigate cross-modal discriminators and cycle losses for a more robust network optimization.
	\item Extensive results on four challenging guided image-to-image translation tasks, i.e., person image generation, facial expression generation, hand gesture-to-gesture translation, and cross-view image translation demonstrate the effectiveness of the proposed C2GAN and show more photorealistic images compared with state-of-the-art models. 
\end{itemize}

Part of this work has been published in \cite{tang2019cycle}. 
We extend it in numerous ways: 
1) A more detailed analysis is presented in ``Introduction'' section by giving a deeper analysis on the motivation and the difference from relevant works.
2) We extend the model proposed in \cite{tang2019cycle} to a unified GAN framework for handling different guided image-to-image translation tasks.
3) We present an in-depth description of the proposed approach, providing all the architectural and implementation details of the method, with special emphasis on guaranteeing the reproducibility of the experiments. 
4) We substantially extend the experimental evaluation. 
\section{Related Work}
\label{sec_rel}

\noindent \textbf{Generative Adversarial Networks (GANs)} have shown the capability of generating high-quality images. To generate images that meet user requirements, Conditional GAN (CGAN)~\cite{mirza2014conditional} is employing the conditioned guidance information to guide the image generation process. A CGAN model always combines a vanilla GAN and an external source, such as segmentation maps \cite{tang2020dual,regmi2019cross,tang2019local}, conditional images~\cite{isola2016image}, and attention maps~\cite{tang2019attentiongan,tang2019attention}. However, synthesizing images based on global constraints does not provide control over pose, object location, or shape.

\noindent \textbf{Image-to-Image Translation} models use input-output data to learn a mapping between the source domain and the target domain. 
Isola et al. propose Pix2pix~\cite{isola2016image}, which employs a CGAN to learn a image mapping from the input domain to the output domain. 
Moreover, unpaired image-to-image translation approaches \cite{tang2018dual,yi2017dualgan,tang2019attention,tang2019attentiongan} have been proposed to learn the mapping without paired training data. 
However, these existing image translation models are inefficient and ineffective as indicated in the introduction section. Most importantly, these aforementioned approaches cannot handle some specific guided image-to-image translation tasks, such as person image generation~\cite{siarohin2017deformable}, and hand gesture-to-gesture translation~\cite{tang2018gesturegan}.

\noindent \textbf{Guided Image-to-Image Translation.}
To address these aforementioned limitations, several works have been proposed to generate images based on object keypoints \cite{ma2017pose,wang2018every}, human hand/body skeleton \cite{tang2020bipartite,yan2017skeleton,tang2020xinggan}, and scene segmentation map \cite{regmi2019cross,liu2020exocentric,tang2019local}. 
For instance, Wang et al. \cite{wang2018every} propose a
Conditional MultiMode Network (CMM-Net) for facial landmark guided smile generation.
Tang et al. \cite{tang2018gesturegan} propose a novel GestureGAN to perform the hand gesture-to-gesture translation task conditioned on an input image and several novel hand skeletons.
Park et al. \cite{park2019semantic} propose a novel spatially-adaptive normalization for the semantic image synthesis task based on semantic labels.
These methods only focus on a single image generation task.

We propose a multitask framework aiming at handling two tasks using a single network, i.e., image generation and guidance generation.
During the training stage, the two generation tasks are restricted mutually by the proposed three cycles and then benefit from each other.
To the best of our knowledge, the proposed C2GAN is the first attempt to generate both the image and the guidance domain in an interactive generation strategy within a unified cycle in cycle GAN model, for guided image-to-image translation tasks. 
\section{Cycle In Cycle GAN (C2GAN)} 
\label{formulation}

\subsection{Model Overview}
The proposed C2GAN learns two different generators in a single network, i.e., image generator and guidance generator.
The two generators are mutually connected through three novel generative adversarial cycles, i.e., one image-oriented cycle and two guidance-oriented cycles. In the training stage, these three cycles are jointly optimized in an end-to-end way and each generator can benefit from the others due to the richer cross-modal information and the crossing cycle supervision. The core framework of C2GAN is illustrated in Fig.~\ref{fig:excyclegan}. 

\subsection{Image-Domain Generative Adversarial Cycle}
\noindent\textbf{I2I2I Cycle.} 
The image cycle I2I2I aims to generate the image $I_y^{'}$ by using the combination of the input image $I_x$ and the target guidance $L_y$,
and then reconstruct the input image $I_x$ by using the combination of the generated image $I_y^{'}$ and the guidance $L_x$ of image~$I_x$:
\begin{equation}
[I_x, L_y] \stackrel{G_i}\rightarrow [I_y^{'}, L_x] \stackrel{G_i}\rightarrow I_x^{'},
\end{equation}
where $G_i$ is the image generator.

Different from the previous guided image-to-image translation methods such as PG2~\cite{ma2017pose}, X-Fork \cite{regmi2018cross}, and PoseGAN~\cite{siarohin2017deformable} employing only one mapping $[I_x, L_y] \stackrel{G_i} \rightarrow I_y^{'}$, StarGAN~\cite{choi2017stargan} employs the target and the original domain labels $l_y$ and $l_x$ as extra guidance information to reconstruct the input image. However, StarGAN can only handle tasks which have a fixed number of the image categories. 
To solve this limitation, we replace the domain labels $l_y$ and $l_x$ in StarGAN by using the guidances $L_y$ and $L_x$. The guidance can be object keypoints, human skeletons, or scene segmentation maps. Specifically, $I_x$ and $L_y$ are first fed into the image generator $G_i$ to generate the desired image $I_y^{'}$.
Next, the generated image $I_y^{'}$ and the guidance $L_x$ are concatenated as the input of $G_i$ to reconstruct the original image $I_x$.
In this way, the forward and backward consistency can be guaranteed.

\noindent \textbf{Image Generator.}
The U-Net architecture~\cite{ronneberger2015u} is adopted for our image generator $G_i$.
U-Net consists of an encoder and a decoder with skip connections between them. 
The generator $G_i$ is used two times for generating image $I_y^{'}$ and reconstructing the original image $I_x^{'}$. To reduce the model capacity, the image generator $G_i$ shares parameters between image generation and reconstruction. For image generation, the target of $G_i$ is generating an image $I_y^{'}{=}G_i(I_x, L_y)$ conditioned on the target guidance $L_y$ which is similar to the real image $I_y$. For image reconstruction, the goal of $G_i$ is recovering an image $I_x^{'}{=}G_i(I_y^{'}, L_x)$ that looks similar to the input image $I_x$. The image generator $G_i$ learns a combined data distribution between the image generation and the image reconstruction by sharing parameters, meaning that $G_i$ receives double data during the network optimization compared to those generators without using the parameter-sharing strategy.

\noindent \textbf{Cross-Modal Image Discriminator.}
Different from previous works such as PG2~\cite{ma2017pose} employing a single-modal discriminator, we propose a novel cross-modal discriminator which receives both image and guidance data as input (Fig.~\ref{fig:excyclegan}). 
The image discriminator $D_i$ receives two images and one guidance data as input.
More specifically, $D_i$ aims to distinguish between the generated triplet $[I_x, L_y, G_i(I_x, L_y)]$ and the real triplet $[I_x, L_y, I_y]$ during the image generation stage.

We further propose an image adversarial loss based on the vanilla adversarial loss \cite{goodfellow2014generative}, which can be expressed as:
\begin{equation}
\begin{aligned}
& \mathcal{L}^i_{GAN}(G_i, D_i, I_x, I_y, L_y) \\ 
= & \mathbb{E}_{I_x, L_y, I_y\sim{p_{\rm data}}(I_x, L_y, I_y)}\left[ \log D_i([I_x, L_y, I_y])\right]  \\ 
+ &  \mathbb{E}_{I_x, L_y\sim{p_{\rm data}}(I_x, L_y)}[\log (1\!\!-\!\!D_i([I_x, L_y, G_i(I_x, L_y)]))],
\end{aligned}
\label{equ:cgan1}
\end{equation}
where the image generator $G_i$ tries to minimize the image adversarial loss $\mathcal{L}^i_{GAN}(G_i, D_i, I_x, I_y, L_y)$ while the image discriminator $D_i$ tries to maximize it.

Another image adversarial loss for the image reconstruction mapping $G_i: [I_y^{'}, L_x] \rightarrow I_x^{'}$ is defined as:
\begin{equation}
\begin{aligned} 
& \mathcal{L}^i_{GAN}(G_i, D_i, I_x, I_y, L_x) \\
=  & \mathbb{E}_{I_x, L_x, I_y\sim{p_{\rm data}}(I_x, L_x, I_y)}[\log D_i([I_y, L_x, I_x])]  \\
+ & \mathbb{E}_{I_y^{'}, L_x, I_y\sim{p_{\rm data}}(I_y^{'}, L_x, I_y)}[\log (1\!\!-\!\! D_i([I_y, L_x, G_i(I_y^{'}, L_x)]))],
\end{aligned}
\label{equ:cgan2}
\end{equation}
where the image discriminator $D_i$ aims at distinguishing between the fake triplet~$[I_y, L_x, G_i(I_y^{'}, L_x)]$ and the real triplet $[I_y, L_x, I_x]$. Therefore, the overall image adversarial loss is the sum of Eq.~\eqref{equ:cgan1} and Eq.~\eqref{equ:cgan2}:
\begin{equation}
\begin{aligned} 
& \mathcal{L}^i_{GAN}(G_i, D_i, I_x, I_y, L_x, L_y)  \\ 
= & \mathcal{L}^i_{GAN}(G_i, D_i, I_x, I_y, L_y) 
+   \mathcal{L}^i_{GAN}(G_i, D_i, I_x, I_y, L_x).
\end{aligned}
\end{equation}

\begin{SCfigure}[1][!t]
	\centering
	\includegraphics[scale=0.25]{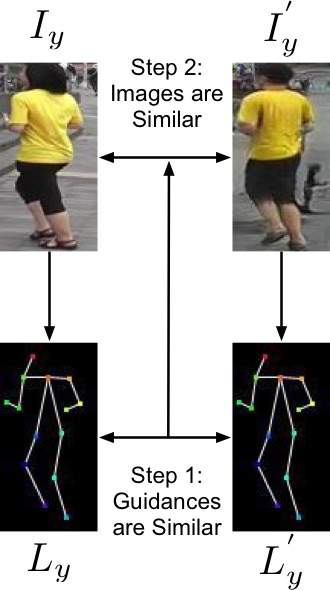}
	\caption{The motivation of the guidance cycle. If the generated guidance $L_y^{'}$ is close to the real guidance $L_y$, then the corresponding images (i.e., $I_y^{'}$ and $I_y$) should be similar.}
	\label{fig:guidance}
	\vspace{-0.4cm}
\end{SCfigure}

\noindent \textbf{Image Cycle-Consistency (IC) Loss.}
We also propose the IC loss to better learn the image cycle I2I2I:
\begin{equation}
\begin{aligned}
&  \mathcal{L}^i_{CYC}(G_i, I_x, L_x, L_y) \\
=  & \mathbb{E}_{I_x, L_x, L_y\sim{p_{\rm data}}(I_x, L_x, L_y)}[\Arrowvert G_i(G_i(I_x, L_y), L_x)-I_x\Arrowvert_1].
\end{aligned}
\label{equ:cycleganloss}
\end{equation}
The reconstructed image $I_x^{'}{=}G_i(G_i(I_x, L_y), L_x)$ should closely match with the input image $I_x$.
Notably, the image generator $G_i$ is used two times with the parameter-sharing strategy and the $L1$ distance is adopted in Eq.~\eqref{equ:cycleganloss} to compute a pixel-to-pixel difference between the recovered image $I_x^{'}$ and the real input image $I_x$.

\subsection{Guidance-Domain Generative Adversarial Cycle}
The motivation of the guidance cycle is that, if the generated guidance is similar to the real guidance, then the corresponding two images should be very close (see Fig.~\ref{fig:guidance}).
The proposed C2GAN has two guidance cycles, i.e., G2I2G and G2R2G, as shown in Fig.~\ref{fig:excyclegan}. 
Both cycles can provide extra supervision information for better optimizing the image cycle I2I2I.

\noindent\textbf{G2I2G Cycle.}
For the G2I2G cycle, $[I_x, L_y]$ is first fed into the image generator $G_i$ to produce the target image $I_y^{'}$.
Then the guidance generator $G_g$ tries to produce the guidance $L_y^{'}$ from the generated image $I_y^{'}$. The generated guidance $L_y^{'}$ should be very close to the real guidance $L_y$. The formulation of the G2I2G cycle can be expressed as:
\begin{equation}
\begin{aligned}
& [I_x, L_y] \stackrel{G_i}\rightarrow I_y^{'} \stackrel{G_g}\rightarrow L_y^{'}.
\end{aligned}
\end{equation}

\noindent\textbf{G2R2G Cycle.}
For the G2R2G cycle, the generated image $I_y^{'}$ and guidance $L_x$ are first concatenated, and then fed into $G_i$ to produce the recovered image $I_x^{'}$. 
Next, the guidance generator $G_g$ generates the guidance $L_x^{'}$ from the recovered image $I_x^{'}$. 
We assume that the generated guidance $L_x^{'}$ is very similar to the real guidance $L_x$.
The G2R2G cycle can be formulated as:
\begin{equation}
\begin{aligned}
& [I_y^{'}, L_x] \stackrel{G_i}\rightarrow I_x^{'} \stackrel{G_g}\rightarrow L_x^{'}.
\end{aligned}
\end{equation}
Both generated guidances $L_y^{'}{=} G_g(G_i(I_x, L_y))$ and $L_x^{'} {=} G_g(G_i(I_y^{'}, L_x))$ should have a close match to the real guidance $L_y$ and $L_x$.
Note that the guidance generator $G_g$ could share parameters between these two cycles.

\noindent\textbf{Guidance Generator.}
The U-Net structure~\cite{ronneberger2015u} is employed for our guidance generator $G_g$.
The input of $G_g$ is an image and the output is a guidance.
The guidance generator respectively produces $L_y^{'}{=}G_g(I_y^{'})$ and $L_x^{'}{=}G_g(I_x^{'})$ from the generated images $I_y^{'}$ and $I_x^{'}$, which further provide more supervision gradient to guide the image generator $G_i$ to produce more realistic images. 

\noindent\textbf{Cross-Modal Guidance Discriminator.}
As shown in Fig.~\ref{fig:excyclegan}, the proposed guidance discriminator $D_g$ is a cross-modal discriminator receiving both image and guidance data as inputs.
Thus, the guidance adversarial loss for $D_g$ can be defined as:
\begin{equation}
\begin{aligned}
& \mathcal{L}^g_{GAN}(G_g, D_g, I_y^{'}, L_y)  \\
= &  \mathbb{E}_{I_y^{'}, L_y\sim{p_{\rm data}}(I_y^{'}, L_y)}\left[ \log D_g([I_y^{'}, L_y])\right]  \\ 
+ & \mathbb{E}_{I_y^{'}\sim{p_{\rm data}}(I_y^{'})}[\log (1 - D_g([I_y^{'}, G_g(I_y^{'})]))],
\end{aligned}
\label{equ:lcgan1}
\end{equation}
where the guidance generator $G_g$ aims to minimize the guidance adversarial loss $\mathcal{L}^g_{GAN}(G_g, D_g, I_y^{'}, L_y)$ while the guidance discriminator $D_g$ tries to maximize it.
The discriminator $D_g$ aims to distinguish between the fake pair $[I_y^{'}, L_y^{'}]$ and the real pair $[I_y^{'}, L_y]$.

A similar guidance adversarial loss for the mapping function $G_g: I_x^{'} \rightarrow L_x^{'}$ is defined as:
\begin{equation}
\begin{aligned} 
& \mathcal{L}^g_{GAN}(G_g, D_g, I_x^{'}, L_x)   \\
= & \mathbb{E}_{I_x^{'}, L_x\sim{p_{\rm data}}(I_x^{'}, L_x)}\left[ \log D_g([I_x^{'}, L_x])\right]  \\
+ & \mathbb{E}_{I_x^{'}\sim{p_{\rm data}}(I_x^{'})}[\log (1 - D_g([I_x^{'}, G_g(I_x^{'})]))],
\end{aligned}
\label{equ:lcgan2}
\end{equation}
where the guidance discriminator $D_g$ aims to distinguish between the fake pair $[I_x^{'}, L_x^{'}]$ and the real pair $[I_x^{'}, L_x]$.

Thus, the total guidance adversarial loss is the sum of Eq.~\eqref{equ:lcgan1} and Eq.~\eqref{equ:lcgan2}:
\begin{equation}
\begin{aligned} 
& \mathcal{L}^g_{GAN}(G_g, D_g, I_x^{'}, I_y^{'}, L_x, L_y) \\
= & \mathcal{L}^g_{GAN}(G_g, D_g, I_y^{'}, L_y) 
+ \mathcal{L}^g_{GAN}(G_g, D_g, I_x^{'}, L_x).
\end{aligned}
\end{equation}

\noindent\textbf{Guidance Cycle-Consistency (GC) Loss.}
A novel GC loss is further proposed to better learn both the guidance cycles (i.e., G2I2G and G2R2G), which can be expressed as:
\begin{equation}
\begin{aligned}
& \mathcal{L}^g_{CYC}(G_g, G_i, I_x, I_y^{'}, L_x, L_y)  \\
= & \mathbb{E}_{I_x, L_y\sim{p_{\rm data}}(I_x, L_y)}[\Arrowvert G_g(G_i(I_x, L_y))-L_y\Arrowvert_1] \\
+ &  \mathbb{E}_{I_y^{'}, L_x\sim{p_{\rm data}}(I_y^{'}, L_x)}[\Arrowvert G_g(G_i(I_y^{'}, L_x))-L_x\Arrowvert_1],
\end{aligned}
\label{equ:lcycleganloss}
\end{equation}
where the $L1$ distance is used to compute the pixel-to-pixel difference between the generated guidance (i.e., $L_x^{'}$ and $L_y^{'}$) and the corresponding real guidance (i.e., $L_x$ and $L_y$).

During the training stage, the proposed guidance cycle-consistency loss can back-propagate errors from the guidance generator $G_g$ to the image generator $G_i$ facilitating the optimization of the image generator and then boosting the image generation performance.

\subsection{Joint Optimization Objective}
We follow existing methods \cite{siarohin2017deformable,tang2019multi} and use the image pixel loss to reduce the changes between the generated image $I_y^{'}{=}G_i(I_x, L_y)$ and the corresponding real one $I_y$:
\begin{equation}
\begin{aligned}
& \mathcal{L}^i_{PIXEL}(G_i, I_x, L_y, I_y )   \\
= & \mathbb{E}_{I_x, L_y, I_y \sim{p_{\rm data}}(I_x, L_y, I_y)}[\Arrowvert G_i(I_x, L_y)-I_y\Arrowvert_1],
\end{aligned}
\label{equ:pixelloss}
\end{equation}
where the $L1$ distance is adopted as the loss measurement of the image pixel loss. By doing so, more constrains can be added on both the image generator $G_i$. 

Consequently, the complete objective loss of the proposed C2GAN framework is:
\begin{equation}
\begin{aligned}
& \mathcal{L}(G_i, G_g, D_i, D_g)  \\
= & \lambda^i_{gan} * \mathcal{L}^i_{GAN} + \lambda^i_{cyc} * \mathcal{L}^i_{CYC} + \lambda^i_{pixel} * \mathcal{L}^i_{PIXEL} \\
+  & \lambda^g_{gan} * \mathcal{L}^g_{GAN} + \lambda^g_{cyc} * \mathcal{L}^g_{CYC},
\end{aligned}
\label{eqn:allloss}
\end{equation}
where $\lambda^i_{gan}$, $\lambda^i_{cyc}$, $\lambda^i_{pixel}$, $\lambda^g_{gan}$, and $\lambda^g_{cyc}$ are parameters controlling the relative relation of objectives terms.

\subsection{Implementation Details}
\noindent\textbf{Network Architecture.} 
We adopt the U-Net architecture \cite{ronneberger2015u} consisting of an encoder and a decoder for our generators $G_i$ and $G_g$. 
Moreover, we employ the PatchGAN discriminator~\cite{isola2016image} for our discriminators $D_i$ and $D_g$, which has shown effectiveness in previous image-to-image translation works \cite{isola2016image,zhu2017unpaired}.
The difference between a PatchGAN and a regular GAN discriminator is that the regular GAN maps from an image to a single scalar output, `real' or `fake', whereas the PatchGAN tries to classify if each $N {\times}N$ patch in an image is real or fake.
By doing so, PatchGAN can alleviate the generation of visual artifacts and achieve better performance.

The amount of network parameters in the proposed method is twice that of Pix2pix \cite{isola2016image} because our method contains two generators and two discriminators, while Pix2pix has only one generator and one discriminator.
Although Pix2pix \cite{isola2016image} has fewer network parameters, it is only able to translate between two domains and cannot handle some specific guided image-to-image translation tasks, such as person image generation, facial expression generation, hand gesture-to-gesture translation, and cross-view image translation.
In contrast, our method is a universal method that can handle all these tasks.

\noindent\textbf{Training Strategy.} 
We follow the standard optimization method from~\cite{goodfellow2014generative} to optimize the proposed C2GAN, i.e., we alternate between one gradient descent step on $G_i$, $D_i$, $G_g$, and $D_g$.
The Adam solver~\cite{kingma2014adam}, with a learning rate of 0.0002, and momentum terms $\beta_1{=}0.5$, $\beta_2{=}0.999$, is adopted as our optimizer.
For each task, we keep the same learning rate for the first half of the number of epochs and linearly decay the rate to zero over the next half of the epochs.
Take facial expression generation as an example, we keep the same learning rate for the first 100 epochs and linearly decay the rate to zero over the next 100 epochs.
The proposed C2GAN is trained end-to-end and can generate image and guidance simultaneously, then the generated guidances will benefit the quality of the generated images. Moreover, to slow down the rate of discriminators $D_i$ and $D_g$ relative to generators $G_i$ and $G_g$, we divide the objectives by 2 while optimizing the discriminators.

The public software OpenFace~\cite{amos2016openface} is employed to extract facial landmarks on the Radboud Faces dataset for facial expression generation. 
While OpenPose~\cite{cao2017realtime} is used to extract human hand and body skeleton on the Creative Senz3D and Market-1501 datasets for hand gesture-to-gesture translation and person image generation, respectively.
Next, RefineNet \cite{lin2017refinenet} is employed to extract segmentation maps from the Dayton dataset for cross-view image translation.

\noindent\textbf{Inference Strategy.} 
During the inference stage, the proposed G2GAN receives an image $I_x$ and a guidance $L_y$ into the image generator $G_i$, and outputs a target image $I_y^{'}$. At the same time, the guidance generator $G_g$ receives the image $I_x$ as input and outputs the corresponding guidance $L_x^{'}$.

\noindent\textbf{Parameter Setting.}
For a fair comparison, all competing models are trained for 200 epochs on the Radboud Faces dataset for facial expression generation.
All models are trained around 90 epochs on person image generation.
For hand gesture-to-gesture translation, all models are trained with 20 epochs.
For cross-view image translation, we train the models for 35 epochs. 
The mask loss proposed in PG2~\cite{ma2017pose} is also used for person image generation.
Our C2GAN is implemented using the public deep learning software PyTorch. 
\section{Experiments}
\label{experiment}

\subsection{Person Image Generation}
\noindent\textbf{Datasets.} 
We employ the Market-1501 dataset~\cite{zheng2015scalable} for the person image generation task. This dataset~\cite{zheng2015scalable} is a challenging person-reID dataset containing 32,668 images of 1,501 persons collected from six surveillance cameras. We adopt the training and testing splits used in \cite{siarohin2017deformable} and obtain 263,631 and 12,000 pairs for the training and testing subset.

\begin{figure}[!t] \small
	\centering
	\includegraphics[width=1\linewidth]{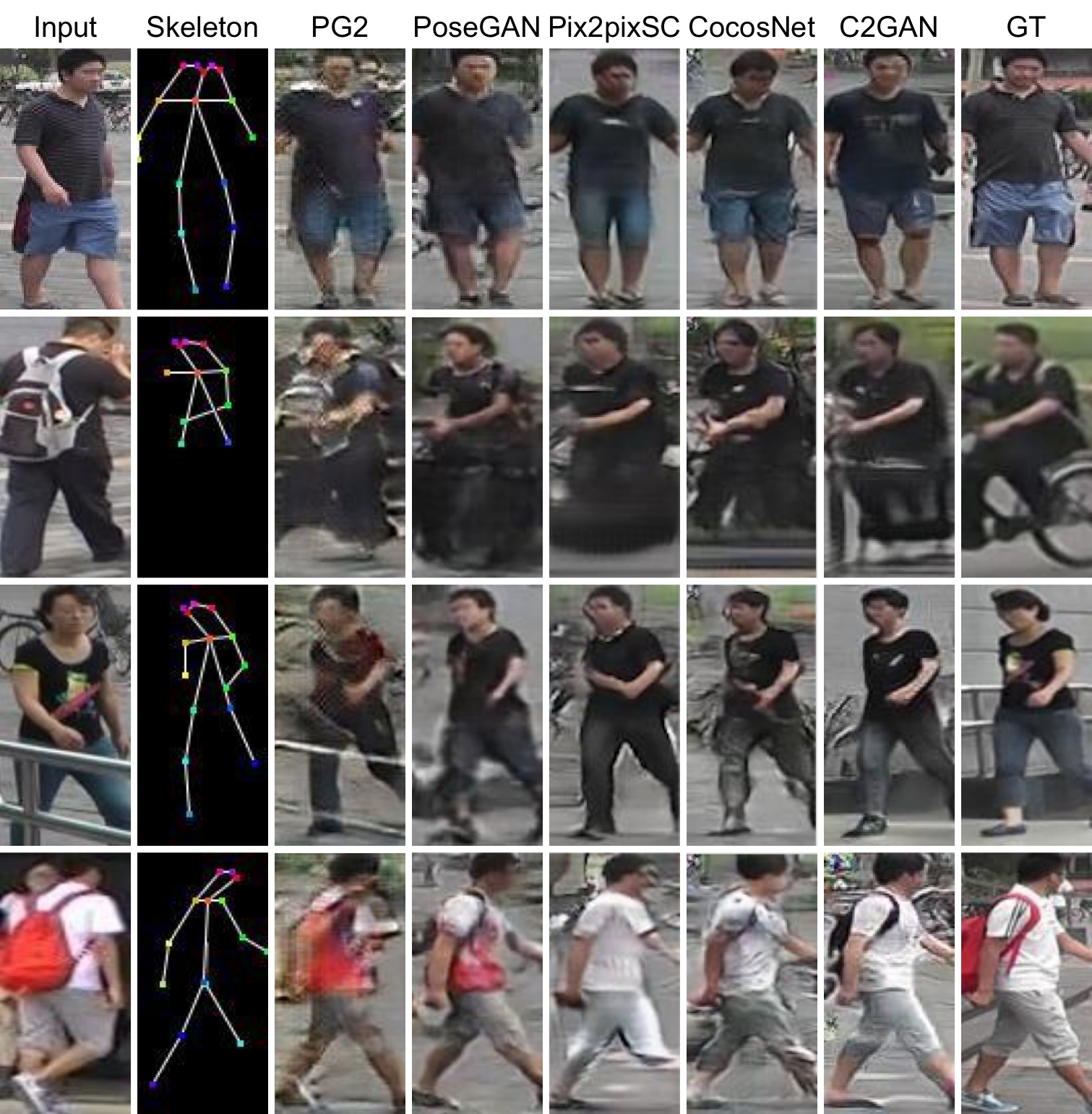}
	\caption{Qualitative comparison of person image generation on the Market-1501 dataset. From left to right: Input, Body Skeleton, PG2~\cite{ma2017pose}, PoseGAN~\cite{siarohin2017deformable}, Pix2pixSC \cite{wang2019example}, CocosNet \cite{zhang2020cross}, C2GAN (Ours), and Ground Truth (GT).}
	\label{fig:result_pose}
	\vspace{-0.4cm}
\end{figure}

\begin{figure*}[!t] \small
	\centering
	\includegraphics[width=1\linewidth]{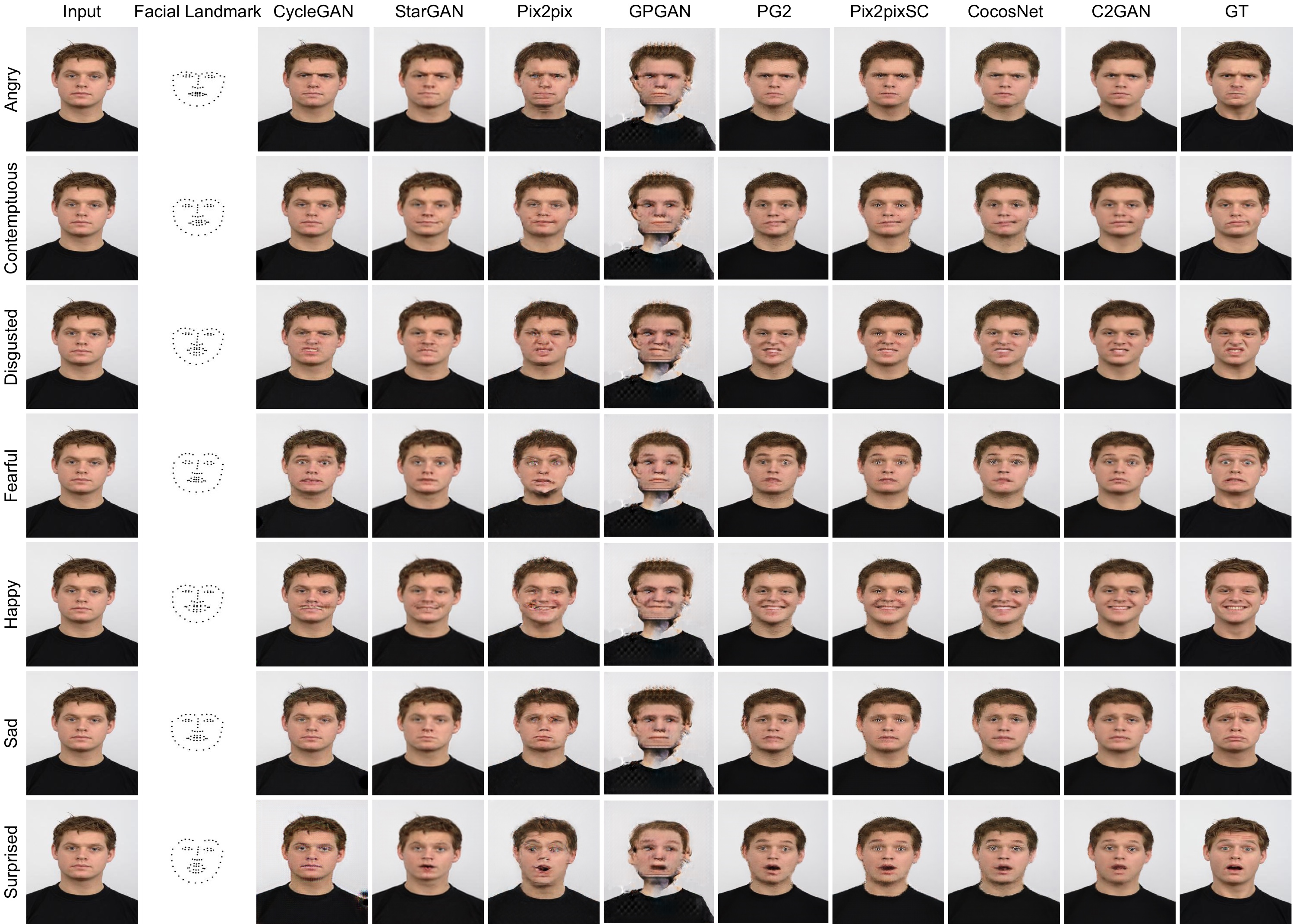}
	\caption{Qualitative comparison of facial expression generation on the Radboud Faces dataset. From left to right: Input, Facial Landmark, CycleGAN \cite{zhu2017unpaired}, StarGAN~\cite{choi2017stargan}, Pix2pix~\cite{isola2016image}, GPGAN~\cite{di2017gp}, PG2~\cite{ma2017pose}, Pix2pixSC \cite{wang2019example}, CocosNet \cite{zhang2020cross}, C2GAN (Ours), and Ground Truth (GT).}
	\label{fig:face}
	\vspace{-0.4cm}
\end{figure*}

\begin{table*}[!t] \small
	\centering
	\caption{Quantitative comparison of person image generation on the Market-1501 dataset. For all the metrics, higher is better.}
	\begin{tabular}{rcccccc} \toprule
		Model &  AMT (R2G) $\uparrow$     & AMT (G2R) $\uparrow$    & SSIM $\uparrow$          & IS  $\uparrow$           & Mask-SSIM  $\uparrow$    & Mask-IS $\uparrow$ \\ \midrule 	
		PG2~\cite{ma2017pose}                  &  11.2          & 5.5           & 0.253          & 3.460          & 0.792          & 3.435   \\ 
		DPIG~\cite{ma2018disentangled}               &  -             & -             & 0.099          & \textbf{3.483} & 0.614          & 3.491   \\  
		PoseGAN~\cite{siarohin2017deformable}  &  22.7        & \textbf{50.2}  &\textbf{0.290} & 3.185          & 0.805          & 3.502   \\ 
		Pix2pixSC \cite{wang2019example} & {18.6} & {41.5} & {0.275} & {3.141} & {0.790} & {3.468} \\
		CocosNet \cite{zhang2020cross} & {20.1} & {45.7} & {0.280} &{3.275} & {0.801} & {3.514} \\
		C2GAN (Ours)                   &  \textbf{23.8} & 47.3          & 0.285          & 3.362          & \textbf{0.813} & \textbf{3.526}   \\ \hline
		Real Data & -              & -             & 1.000          & 3.860          & 1.000          & 3.360  \\ \bottomrule 
	\end{tabular}
	\label{tab:result_market}
	\vspace{-0.4cm}
\end{table*} 

\begin{figure*}[!t] \small
	\centering
	\includegraphics[width=0.8\linewidth]{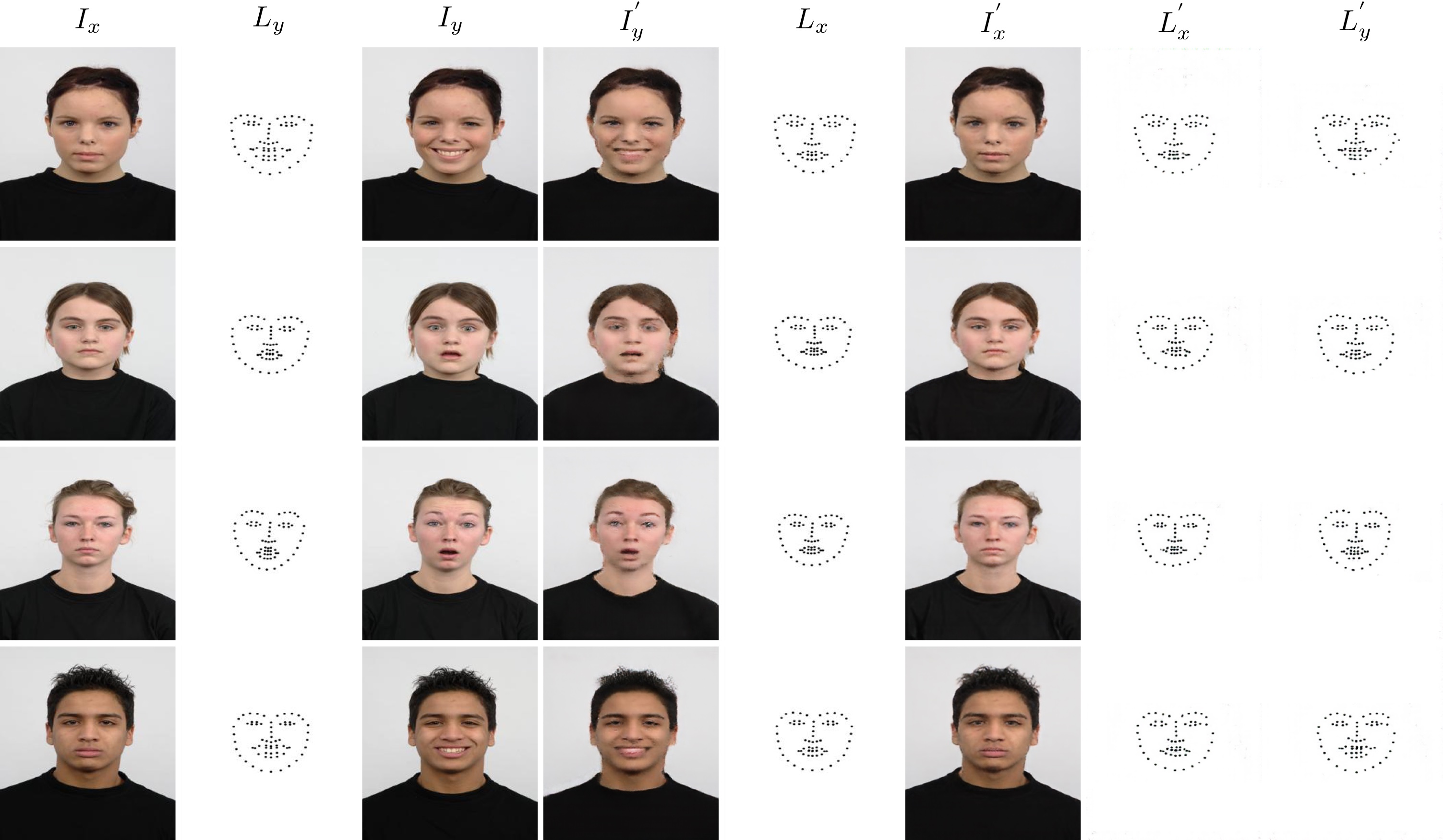}
	\caption{Visualization of facial landmark generation on the  facial expression generation task.}
	\label{fig:keypoint_visualization}
	\vspace{-0.4cm}
\end{figure*}

\noindent\textbf{Evaluation Metrics.} 
We follow \cite{siarohin2017deformable,ma2017pose} and adopt Inception Score (IS)~\cite{salimans2016improved}, Structural Similarity (SSIM)~\cite{wang2004image}, and their masked versions Mask-SSIM and Mask-IS as our evaluation metrics. Moreover, the AMT perceptual user study is adopted to evaluate the generated images by different models.

\noindent \textbf{State-of-the-Art Comparison.} We compare C2GAN with  PG2~\cite{ma2017pose}, DPIG~\cite{ma2018disentangled}, PoseGAN~\cite{siarohin2017deformable}, Pix2pixSC \cite{wang2019example}, and CocosNet \cite{zhang2020cross}. Different from these models which focus on person image generation, our method is a general framework and learns image and guidance generation simultaneously in a joint network. Quantitative results are shown in Table \ref{tab:result_market}.
C2GAN achieves better results than PG2, DPIG, Pix2pixSC, and CocosNet.
Moreover, compared to PoseGAN \cite{siarohin2017deformable}, C2GAN yields  better results on most metrics, i.e., AMT (R2G), IS, mask-SSIM, and mask-IS.  
Qualitative comparison results compared with PG2 and PoseGAN are shown in Fig.~\ref{fig:result_pose}. C2GAN can generate more clear and visually plausible person images than both leading methods, validating the effectiveness C2GAN. 
Moreover, our generated images are more similar to the ground truth.

\subsection{Facial Expression Generation}
\noindent\textbf{Datasets.} 
We employ the Radboud Faces dataset~\cite{langner2010presentation} for the facial expression generation task.
This dataset contains over 8,000 color face images with eight different facial expressions.
We randomly select 67\% of the images for training and the rest 33\% images for testing. We remove the images in which the face is not correctly detected by OpenFace~\cite{amos2016openface}, then combine two different facial expression images of the same person to form an image pair for training. Therefore, 5,628 and 1,407 image pairs are obtained for training and testing, respectively.

\begin{table}[!t] \small
	\centering
	\caption{Quantitative comparison of facial expression generation on the Radboud Faces dataset. For all the metrics except LPIPS, higher is better.} 
	\begin{tabular}{rcccc} \toprule
		Model                          & AMT $\uparrow$  & SSIM $\uparrow$  & PSNR $\uparrow$   & LPIPS $\downarrow$ \\ \midrule
		CycleGAN \cite{zhu2017unpaired} & 19.5 & 0.8307 & 18.8067 & - \\
		StarGAN~\cite{choi2017stargan} & 24.7            & 0.8345           & 19.6451           & - \\
		Pix2pix~\cite{isola2016image}  & 13.4            & 0.8217           & 19.9971           & 0.1334 \\ 
		GPGAN~\cite{di2017gp}          & 0.3             & 0.8185           & 18.7211           & 0.2531 \\ 
		PG2~\cite{ma2017pose}          & 28.4            & 0.8462           & 20.1462           & 0.1130 \\  
		Pix2pixSC \cite{wang2019example} & 30.8 & 0.8433 & 20.3584 & 0.1042 \\
		CocosNet \cite{zhang2020cross} & 31.3 & 0.8524 & 20.7915 & 0.0985 \\
		C2GAN (Ours)  & \textbf{34.2}          & \textbf{0.8618}           & \textbf{21.9192}           & \textbf{0.0934} \\   \bottomrule	
	\end{tabular}
	\label{tab:result_face}
 \vspace{-0.4cm}
\end{table}

\noindent\textbf{Evaluation Metrics.} 
We first adopt the AMT user study to evaluate the generated images.
Moreover, we employ SSIM~\cite{wang2004image}, Peak Signal-to-Noise Ratio
(PSNR), and LPIPS \cite{zhang2018unreasonable} for quantitative measurements.
SSIM and PSNR measure the image quality from a pixel level, while LPIPS evaluates the generated images from a deep feature level. 

\noindent \textbf{State-of-the-Art Comparisons.}
The proposed method is compared with several facial image generation models, i.e., CycleGAN \cite{zhu2017unpaired}, StarGAN~\cite{choi2017stargan}, Pix2pix~\cite{isola2016image}, GPGAN~\cite{di2017gp}, PG2~\cite{ma2017pose}, Pix2pixSC \cite{wang2019example}, and CocosNet \cite{zhang2020cross}.
Note that CycleGAN and StarGAN are unsupervised generation methods, while the others are supervised generation models. The comparisons with CycleGAN and StarGAN are just to see how big the gap between supervised and unsupervised methods is for this task.
The results are shown in Table \ref{tab:result_face}.
We observe that the proposed C2GAN achieves the best results on all four evaluation metrics, validating the effectiveness of our method.

\begin{figure*}[!t] \small
	\centering
	\includegraphics[width=1\linewidth]{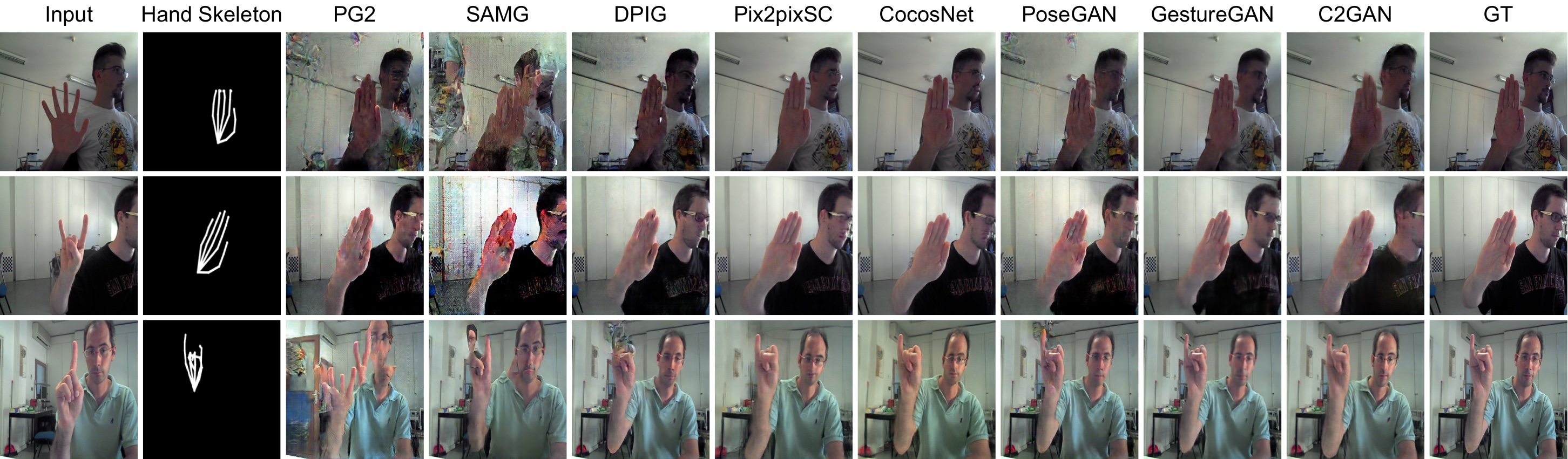}
	\caption{Qualitative comparison of hand gesture-to-gesture translation on the Senz3d dataset. From left to right: Input, Hand Skeleton, PG2~\cite{ma2017pose}, SAMG~\cite{yan2017skeleton}, DPIG~\cite{ma2018disentangled}, Pix2pixSC \cite{wang2019example}, CocosNet \cite{zhang2020cross}, PoseGAN~\cite{siarohin2017deformable}, GestureGAN~\cite{tang2018gesturegan}, C2GAN (Ours), and Ground Truth (GT).}
	\label{fig:hand_results}
	\vspace{-0.4cm}
\end{figure*}

\begin{figure*}[!t] \small
	\centering
	\includegraphics[width=0.8\linewidth]{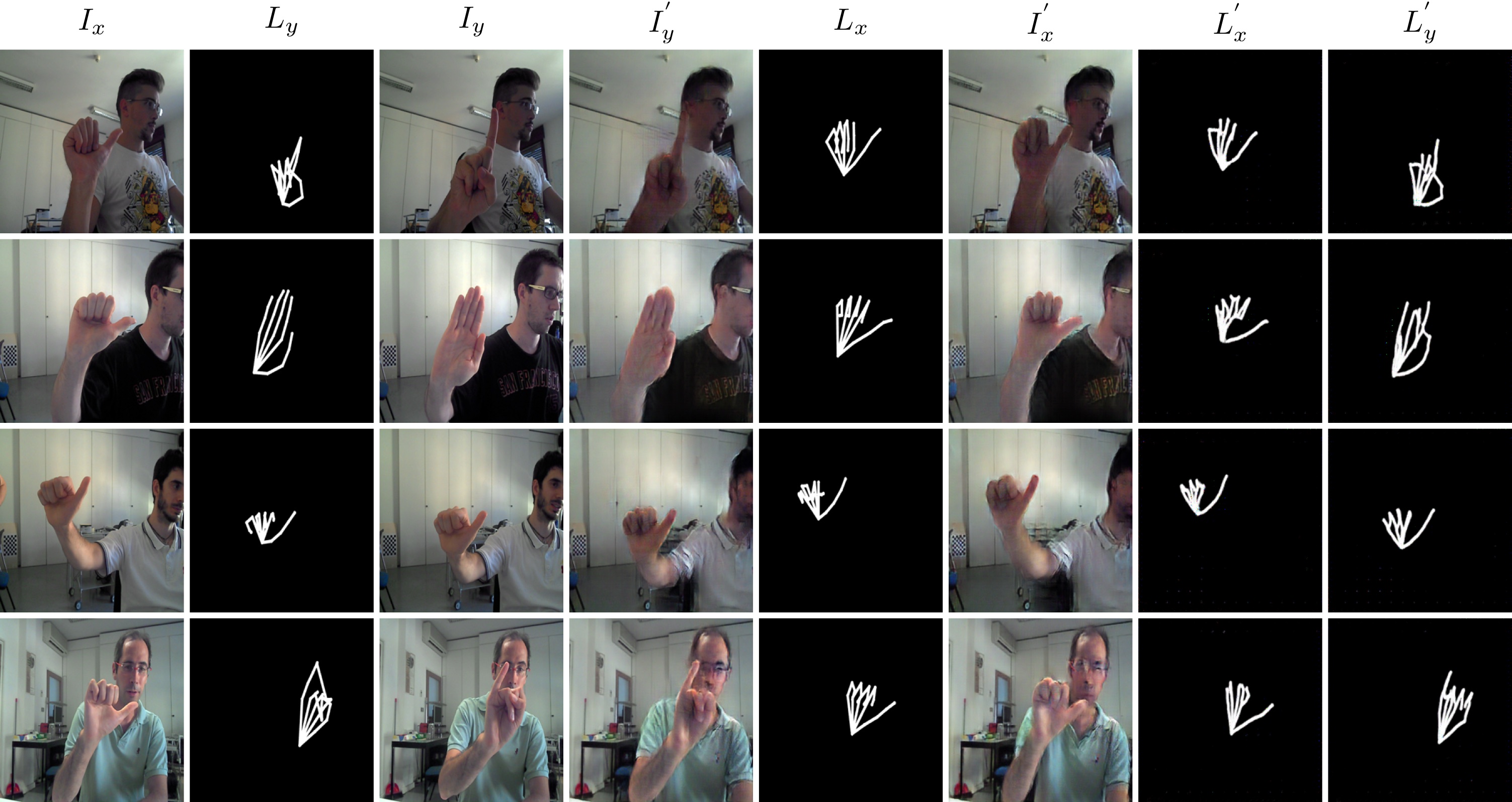}
	\caption{Visualization of hand skeleton generation on the  hand gesture-to-gesture translation task.}
	\label{fig:hand_results_more}
	\vspace{-0.4cm}
\end{figure*}

Qualitative comparison results compared with the leading methods are shown in Fig.~\ref{fig:face}.
Clearly, GPGAN performs worse among all comparison models.
Pix2pix can generate the correct expression, but the faces are distorted.
StarGAN can generate sharper faces, but the details of these generated faces are not convincing.
For instance, the mouths in StarGAN are blurred or not correct.
Moreover, the results of PG2 tend to be blurry. Compared with the existing leading methods, the results generated by the proposed C2GAN are smoother, sharper and contain more details. 
We also show some generated facial landmarks in Fig.~\ref{fig:keypoint_visualization}.
We see that the proposed method not only produces realistic images but also generates reasonable facial landmarks. This is not provided by any existing facial expression generation works.

\subsection{Hand Gesture-to-Gesture Translation}
\noindent\textbf{Datasets.} We follow GestureGAN \cite{tang2018gesturegan} and adopt 
the Creative Senz3D dataset \cite{memo2018head} for the hand gesture-to-gesture translation task.
This dataset contains 11 different hand gestures performed by four people, each performing 
gesture is repeated 30 times, resulting in 4 subjects $\times$
11 gestures $\times$ 30 times = 1320 images in total. 
We follow \cite{tang2018gesturegan} and select 12,800 and 135,504 pairs as testing and training data, respectively.

\begin{table}[!t] \small
	\centering
	\caption{Quantitative comparison of hand gesture-to-gesture translation on the Senz3D dataset. For all metrics except FRD, higher is better.}
		\begin{tabular}{rccc} \toprule
			Method & PSNR $\uparrow$ & AMT $\uparrow$ & FRD $\downarrow$ \\ \midrule	
			PG2~\cite{ma2017pose}                 & 26.5138  & 2.8 & 3.0933 \\ 
			SAMG~\cite{yan2017skeleton}           & 26.9545  & 2.3  & 3.1006\\ 
			DPIG~\cite{ma2018disentangled}        & 26.9451  & 6.9  & 3.0846\\ 
			Pix2pixSC \cite{wang2019example} & {27.0569} & {7.2} & {3.0814} \\
			CocosNet \cite{zhang2020cross} & {27.1532} & {7.9} & {3.0741}\\
		PoseGAN~\cite{siarohin2017deformable} & 27.3014  & 8.6  & 3.0467 \\ 
		GestureGAN~\cite{tang2018gesturegan} & \textbf{27.9749} & \textbf{22.6}  & \textbf{2.9836} \\
			C2GAN (Ours)  & 27.2531 & 12.7 & 3.0573 \\
			\bottomrule		
	\end{tabular}
	\label{tab:gesture_comp}
 \vspace{-0.4cm}
\end{table}

\noindent\textbf{Evaluation Metrics.} 
We follow  \cite{tang2018gesturegan} and adopt Peak Signal-to-Noise Ratio
(PSNR) and FRD \cite{tang2018gesturegan} as evaluation metrics.
PSNR measures the similarity between the real image and the generated image at a pixel level.
FRD measures the distance between the  real image and the fake image from a deep feature level.
Moreover, we follow \cite{tang2018gesturegan} and conduct a user study to evaluate the generated image by different models.

\noindent \textbf{State-of-the-Art Comparisons.}
We adopt the most related several works, i.e., PG2~\cite{ma2017pose}, DPIG~\cite{ma2018disentangled}, PoseGAN~\cite{siarohin2017deformable}, GestureGAN~\cite{tang2018gesturegan}, SAMG~\cite{yan2017skeleton}, Pix2pixSC \cite{wang2019example}, and CocosNet \cite{zhang2020cross}, as our baselines for the facial expression generation task.
Comparison results are shown in Table \ref{tab:gesture_comp}.
The proposed method achieves very competitive results compared with the leading methods.
Specifically, the proposed C2GAN achieves significantly better results than PG2, SAMG, DPIG, Pix2pixSC, and CocosNet on all metrics.
Moreover, PoseGAN obtains slightly better results than our C2GAN on both PSNR and FRD metrics, however, the proposed C2GAN achieves better AMT than PoseGAN.
Moreover, the proposed C2GAN achieves visually better results than PoseGAN, as shown in Fig.~\ref{fig:hand_results}.
Lastly, GestureGAN achieves better results than C2GAN on all metrics.
The reason is that GestureGAN is carefully tailored and designed for the specific hand gesture-to-gesture translation task, meaning that GestureGAN is fine-turned to this task with the network structure, loss objective, and hyper-parameter selection.
However, the proposed C2GAN is a novel and unified GAN model, which can be used to handle all kinds of settings of guided image-to-image translation without modifying the network structure, the loss objective, and hyper-parameters.
Furthermore, our C2GAN can generate both images and guidances, which is not considered in GestureGAN.

\begin{figure*}[!ht] \small
	\centering
	\includegraphics[width=1\linewidth]{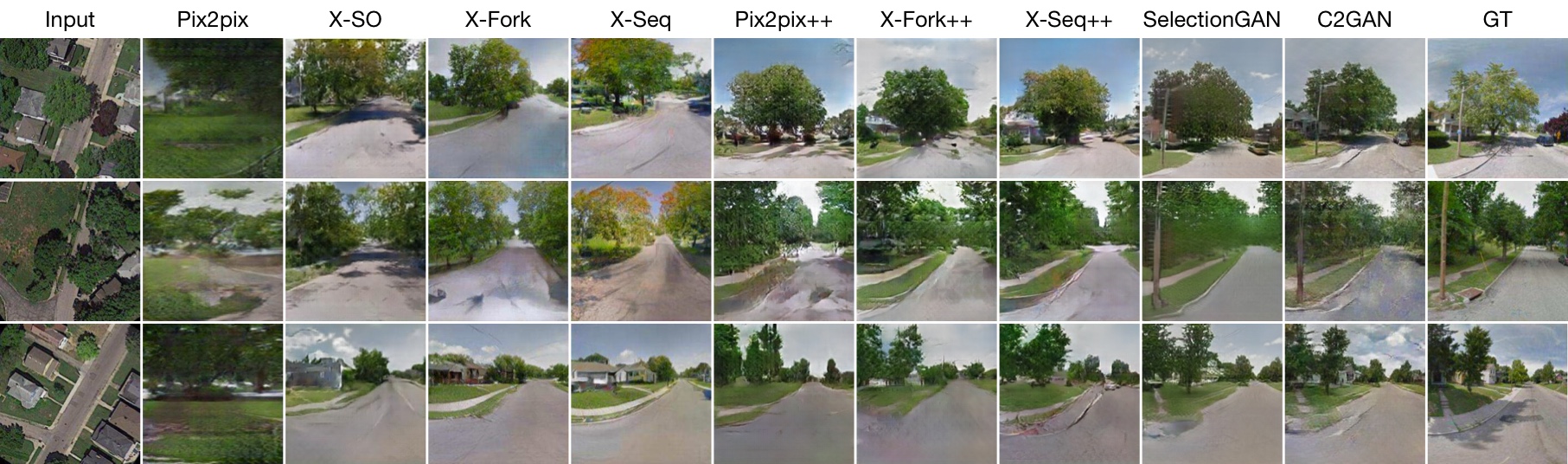}
	\caption{Qualitative comparison of cross-view image translation on the Dayton dataset. From left to right: Input, Pix2pix \cite{isola2016image}, X-SO \cite{regmi2019cross}, X-Fork \cite{regmi2018cross}, X-Seq \cite{regmi2018cross}, Pix2pix++ \cite{isola2016image}, X-Fork++ \cite{regmi2018cross}, X-Seq++ \cite{regmi2018cross}, SelectionGAN \cite{tang2019multi}, C2GAN (Ours), and Ground Truth (GT).}
	\label{fig:result_cross_view}
	\vspace{-0.4cm}
\end{figure*}

\begin{figure*}[!ht] \small
	\centering
	\includegraphics[width=0.8\linewidth]{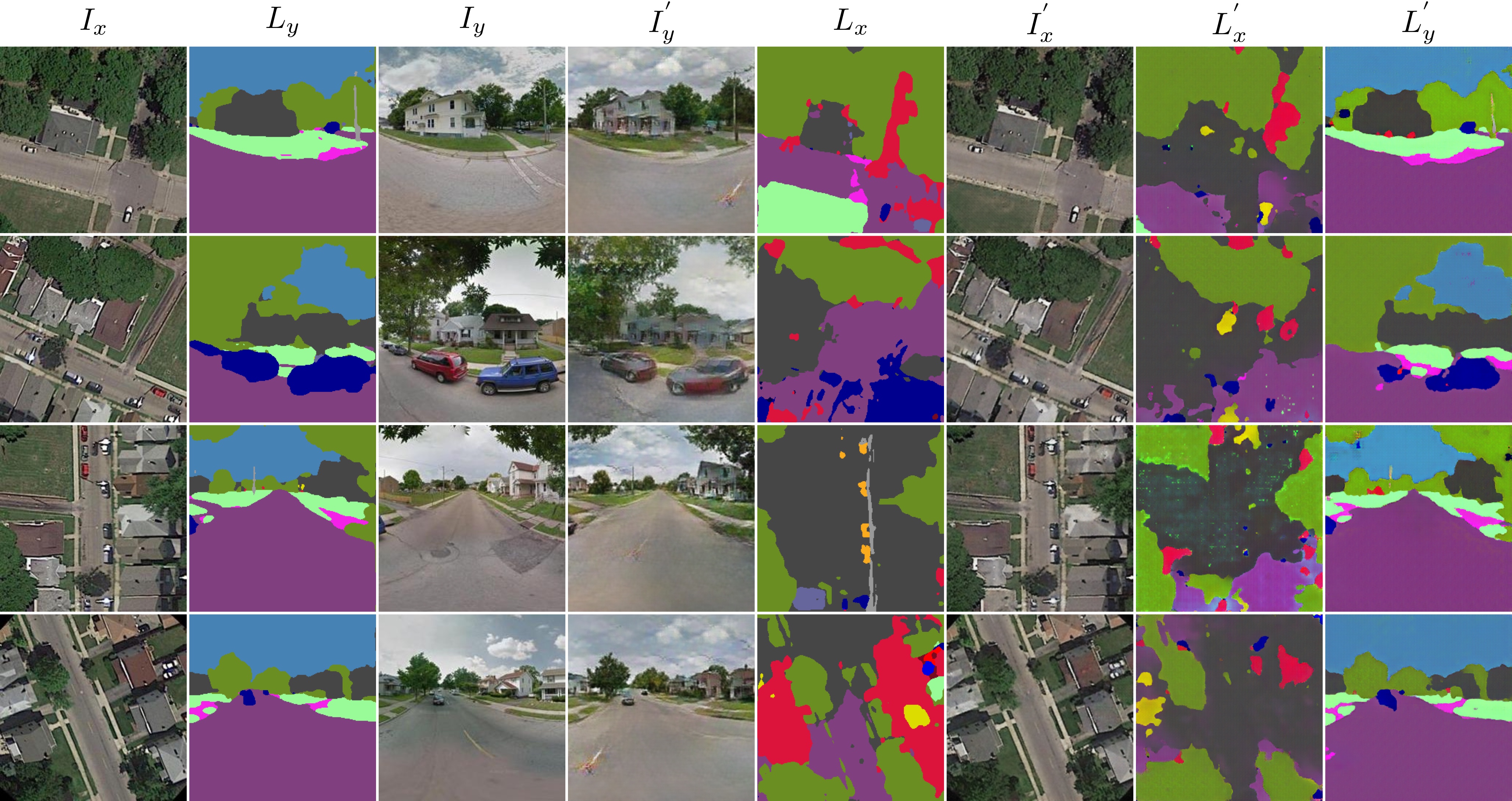}
	\caption{Visualization of segmentation map generation on the  cross-view image translation task.}
	\label{fig:cross_view_results_more}
	\vspace{-0.4cm}
\end{figure*}

\begin{table*}[!t] \small
	\centering
	\caption{Quantitative comparison of cross-view image translation on the Dayton dataset in a2g direction. For all metrics except KL, higher is better.
	}
	\begin{tabular}{rcccccccc} \toprule
		\multirow{2}{*}{Method} & \multicolumn{4}{c}{Accuracy (\%)} & \multicolumn{3}{c}{Inception Score} & \multirow{2}{*}{KL $\downarrow$}  \\ \cmidrule(lr){2-5} \cmidrule(lr){6-8} 
		& \multicolumn{2}{c}{Top-1 $\uparrow$} & \multicolumn{2}{c}{Top-5 $\uparrow$} & All $\uparrow$ & Top-1 $\uparrow$ & Top-5 $\uparrow$ \\ \hline
		Pix2pix \cite{isola2016image}     &6.80  &9.15 &23.55&27.00& 2.8515&1.9342&2.9083 & 38.26 $\pm$ 1.88 \\
		X-SO \cite{regmi2019cross}        &27.56 & 41.15 & 57.96 & 73.20 & 2.9459 & 2.0963 & 2.9980 & 7.20 $\pm$ 1.37 \\
		X-Fork \cite{regmi2018cross}      &30.00 &48.68&61.57&78.84& 3.0720&2.2402&3.0932 &6.00 $\pm$ 1.28 \\
		X-Seq \cite{regmi2018cross}       &30.16 &49.85&62.59&80.70& 2.7384&2.1304&2.7674 & 5.93 $\pm$ 1.32 \\
		Pix2pix++~\cite{isola2016image}   &32.06 &54.70&63.19&81.01&\textbf{3.1709}&2.1200&3.2001& 5.49 $\pm$ 1.25\\
		X-Fork++~\cite{regmi2018cross}    &34.67 & 59.14 &66.37&84.70&3.0737&2.1508&3.0893& 4.59 $\pm$ 1.16 \\
		X-Seq++~\cite{regmi2018cross}     &31.58 & 51.67 &65.21 & 82.48 &3.1703&2.2185&\textbf{3.2444}& 4.94 $\pm$ 1.18 \\
		SelectionGAN~\cite{tang2019multi} & 42.11 & 68.12 & \textbf{77.74} & \textbf{92.89} & 3.0613 & \textbf{2.2707} & 3.1336 & 2.74 $\pm$ 0.86 \\
		C2GAN (Ours) & \textbf{45.80} &\textbf{75.28}& 76.03 & 90.67 & 2.9603 & 2.1225 & 2.9435 & \textbf{2.70 $\pm$ 1.02}  \\ \bottomrule		
	\end{tabular}
	\label{tab:dayton}
	\vspace{-0.4cm}
\end{table*}

Qualitative comparison results compared with PG2, DPIG, PoseGAN, GestureGAN, SAMG are shown in Fig.~\ref{fig:hand_results}.
The proposed method generates much better images than PG2, DPIG, SAMG, and PoseGAN.
Moreover, our results are very close to those generated by GestureGAN. Our C2GAN is a joint learning framework and it is not only able to generate the target images but is also able to produce the hand skeleton of the input image, which will benefit other computer vision tasks such as hand pose estimation.
The results of the generated hand skeletons are shown in Fig.~\ref{fig:hand_results_more}.
The generated hand skeleton $L_x^{'}$ is very similar to the real hand skeleton $L_x$, which verifies the effectiveness of the guidance generator~$G_g$ and our joint learning strategy.

\begin{figure*}[!t] \small
	\centering
	\includegraphics[width=0.8\linewidth]{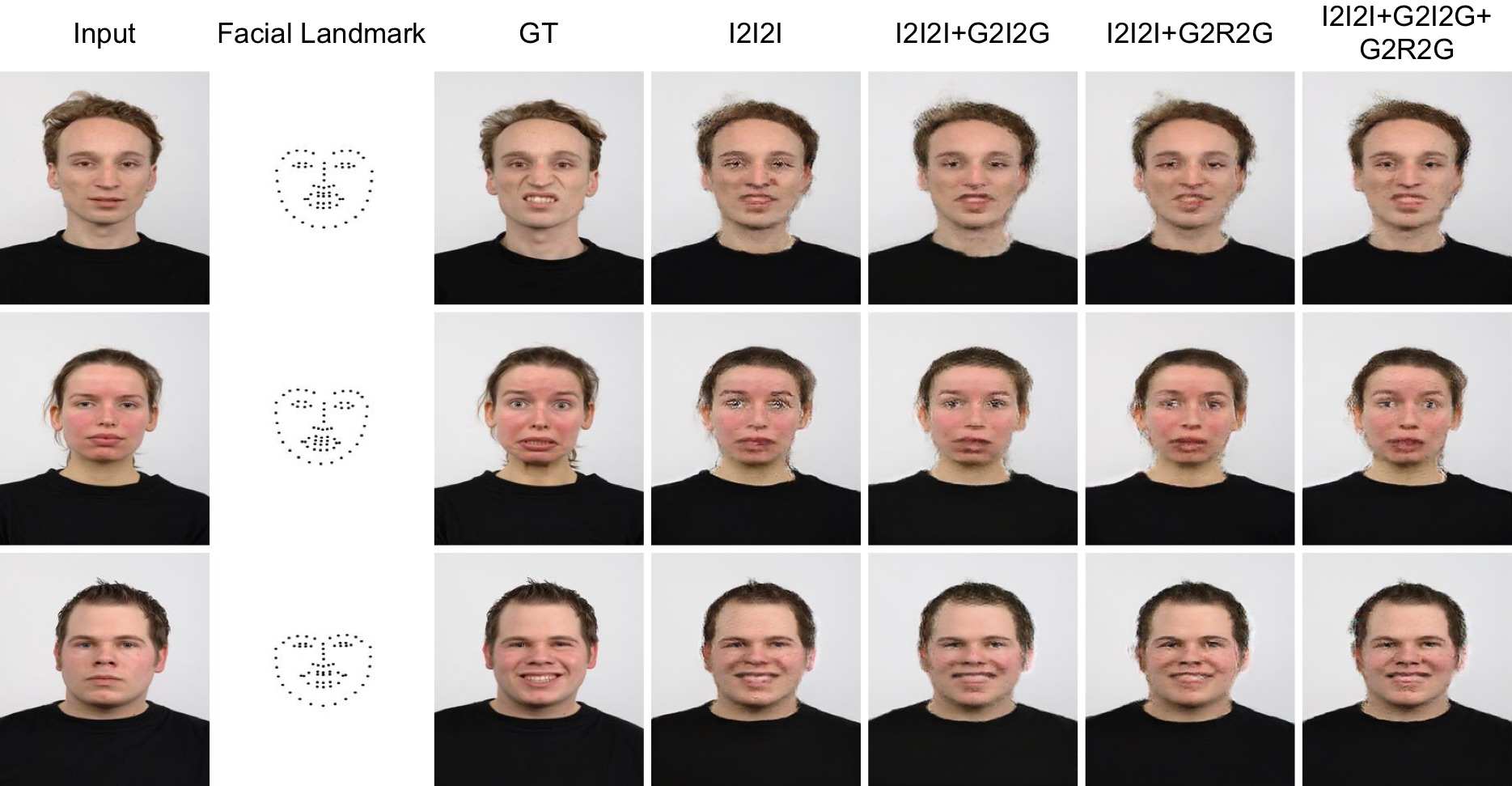}
	\caption{Influence of individual generation cycle on the Radbound Faces dataset.}
	\label{fig:ablation}
	\vspace{-0.4cm}
\end{figure*}

\subsection{Cross-View Image Translation}
\noindent\textbf{Datasets.} 
We follow \cite{regmi2018cross} and adopt the Dayton dataset \cite{vo2016localizing} to evaluate the cross-view image translation task.
This dataset contains 76,048 images and a training/testing split of 55,000/21,048. 
The original size of the image is $354 {\times}354$ resolution.
The images are resized to $256{\times}256$.

\noindent\textbf{Evaluation Metrics.} 
Following \cite{regmi2018cross}, Inception Score (IS), top-k prediction accuracy, and KL score are employed for the quantitative analysis. 
These three metrics evaluate the generated images from a high-level feature space.

\noindent \textbf{State-of-the-Art Comparisons.}
Several leading cross-view image translation methods are adopted as our baselines, i.e., Pix2pix \cite{isola2016image}, X-SO \cite{regmi2019cross}, X-Fork \cite{regmi2018cross}  and X-Seq \cite{regmi2018cross}. 
These methods aim to generate images based on a given image. To further evaluate the proposed C2GAN, we introduce four strong baselines, i.e., Pix2pix++ \cite{isola2016image}, X-Fork++ \cite{regmi2018cross}, X-Seq++ \cite{regmi2018cross} and SelectionGAN \cite{tang2019multi}.
These four models aim to generate images based on a given image and several novel segmentation maps.
Note that we implement Pix2pix++, X-Fork++ and X-Seq++ using their released public code.

Comparison results are shown in Table \ref{tab:dayton}.
The proposed C2GAN achieves the best results on several metrics such as KL and Top-1 Accuracy.
For other metrics, the proposed method still achieves very competitive results, which validates the effectiveness of the proposed C2GAN. 

Several qualitative comparison results are also provided in Fig. \ref{fig:result_cross_view}.
Our C2GAN generates much better realistic images than other baselines. 
Moreover, we show the generated segmentation maps by our method in Fig. \ref{fig:cross_view_results_more}, the proposed C2GAN can generate reasonable segmentation maps, which we believe our method can be used to improve the performance of semantic segmentation tasks.

\begin{table}[!t] \small
	\centering
	\caption{Quantitative comparison of ablation studies on the Radbound Faces dataset. For all metrics, higher is better.} 
	\resizebox{1\linewidth}{!}{%
	\begin{tabular}{lccc}\toprule
		Baseline   &  AMT $\uparrow$ & PSNR $\uparrow$ & SSIM  $\uparrow$ \\ \midrule
		C2GAN w/ I2I2I                      &  25.3 & 21.2030          & 0.8449 \\ 
		C2GAN w/ I2I2I + G2I2G          &   28.2 & 20.8708          & 0.8419 \\ 
		C2GAN w/ I2I2I + G2R2G           & 28.7 & 21.0156          & 0.8437 \\ 
		C2GAN w/ I2I2I + G2I2G + G2R2G & 30.8            & 21.6262          & 0.8540 \\ \hline
		C2GAN w/ Single-Modal $D$      & 26.4 & 21.2794          & 0.8426 \\ \bottomrule
		C2GAN w/ Non-Sharing $G$      & \textbf{32.9}& \textbf{21.6353} & \textbf{0.8611} \\ \bottomrule
	\end{tabular}}
	\label{tab:abl}
	\vspace{-0.4cm}
\end{table}

\subsection{Ablation Study}
We conduct extensive experiments on the Radbound Faces dataset to verify the effectiveness of each component of the proposed C2GAN. 
All experiments are trained with 50 epochs and Table~\ref{tab:abl} shows the quantitative comparison results.

\noindent\textbf{Influence of Individual Generation Cycle.}
To evaluate the influence of individual generation cycle, we test with four different combinations of the cycles, i.e., `I2I2I', `I2I2I+G2I2G', `I2I2I+G2R2G', and `I2I2I+G2I2G+G2R2G'. 
All four combinations use the same training strategies and hyper-parameters. Comparison results are shown in Table~\ref{tab:abl}.
Clearly, `I2I2I', `G2I2G', and `G2R2G' are all critical to the final result and the removal of one of them degrades the generation performance, demonstrating that by using cross-modal data in a joint framework and by making the cycles constraint on each other improve the final generation performance. 
Moreover, `I2I2I+G2I2G+G2R2G' obtains the best performance among all four combination settings.
Meanwhile, `I2I2I+G2I2G+G2R2G' achieves remarkably better results than I2I2I on all metrics, demonstrating the effectiveness of constraining both image and guidance cycles facilitating thus a more robust optimization of the whole model.
Moreover, some visualization results are provided in Fig.~\ref{fig:ablation} to show the influence of each generation cycle. We can obtain the similar conclusion as the one from Table~\ref{tab:abl}, further validating our network design.

\noindent \textbf{Cross-Modal vs. Single-Modal Discriminator.}
We then evaluate the influence of the proposed cross-modal discriminator (`C2GAN w/ I2I2I+G2I2G+G2R2G'). 
Our baseline is the traditional single-modal discriminator (`C2GAN w/ Single-Modal $D$').
The results are listed in Table~\ref{tab:abl}.
The proposed cross-modal discriminator achieves much better results than the single-modal discriminator on all metrics, meaning that the rich cross-modal information helps to learn a better discriminator and thus facilitates the optimization of the generator.

\begin{table}[!t] \small
	\centering
	\caption{Influence of $\lambda^i_{pixel}$.} 
	\begin{tabular}{lccc}\toprule
		SSIM  $\uparrow$  & 1 & 10 & 100  \\ \midrule
		$\lambda^i_{pixel}$ & 0.8143 & \textbf{0.8540} & 0.8349 \\ \bottomrule
	\end{tabular}
	\label{tab:hyper}
	\vspace{-0.4cm}
\end{table}

\noindent \textbf{Parameter Sharing between Generators.}
The parameter sharing could remarkably reduce parameters of the whole network. 
We then evaluate how the parameter-sharing strategy would affect the generation results. 
Specifically, two different baselines are tested: one is `C2GAN w/ I2I2I+G2I2G+G2R2G', which shares the network parameters between the two image generators, and between the two guidance generators, respectively. 
While `C2GAN w/ Non-Sharing $G$' learns four different generators separately. 
As can be seen in Table~\ref{tab:abl}, the non-sharing one achieves slightly better performance than the sharing one. 
However, the number of parameters of non-sharing one is 217.6M, which is twice as much as that of the sharing one. 
This means that the parameter-sharing strategy is a good way to balance both image performance and network overhead.

\noindent \textbf{Influence of Hyper-Parameters.}
For the hyper-parameters in Eq.~\eqref{eqn:allloss}. We first follow Pix2pix \cite{isola2016image} and set the hyper-parameters of adversarial losses (i.e., $\lambda^i_{gan}$ and $\lambda^g_{gan}$) to 1.
Next, we follow CycleGAN \cite{zhu2017unpaired} and set the hyper-parameters of cycle-consistency losses (i.e., $\lambda^i_{cyc}$ and $\lambda^g_{cyc}$) to 10.
Lastly, we investigate the influence of the hyper-parameter of pixel reconstruction loss (i.e., $\lambda^i_{pixel}$) on the performance of our model. 
Comparison results are shown in Table~\ref{tab:hyper}, the proposed method achieves the best performance when $\lambda^i_{pixel}{=}10$.
Therefore, the hyper-parameters $\lambda^i_{gan}$, $\lambda^g_{gan}$, $\lambda^i_{cyc}$, $\lambda^g_{cyc}$, and $\lambda^i_{pixel}$ in Eq.~\eqref{eqn:allloss} are set to 1, 1, 10, 10, and 10, respectively, in all experiments.

\begin{figure}[!t] \small
	\centering
	\includegraphics[width=0.7\linewidth]{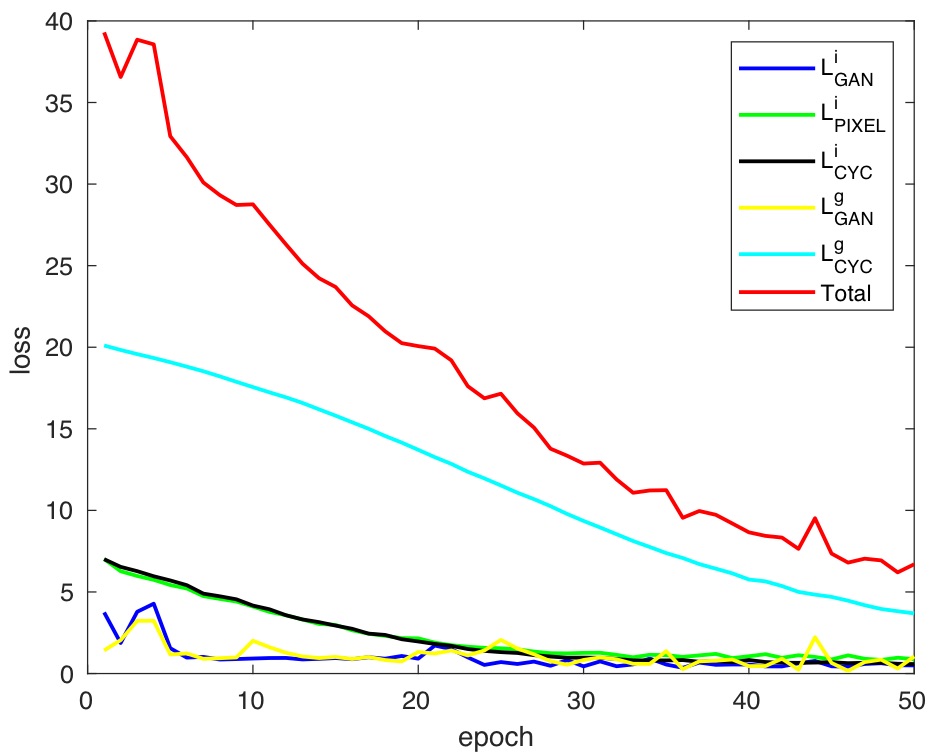}
	\caption{Model convergence loss in Eq.~\eqref{eqn:allloss}.}
	\label{fig:loss}
	\vspace{-0.4cm}
\end{figure}

\noindent \textbf{Model Convergence and Training Time.} 
Fig.~\ref{fig:loss} illustrates the convergence loss of the proposed method in Eq.~\eqref{eqn:allloss}. Note that the proposed model ensures a very fast yet stable convergence.
Moreover, our proposed method takes about 10 hours to finish the training of the ablation study on a single TITAN Xp GUP, while CocosNet \cite{zhang2020cross} and PG2 \cite{ma2017pose} task around 18 and 14 hours, respectively.
\section{Conclusion}
\label{conclusions}

We propose a novel and unified Cycle In Cycle Generative Adversarial Network (C2GAN) for guided image-to-image translation tasks. 
The proposed C2GAN contains two different types of generators, i.e., image-oriented generator and guidance-oriented generator. 
Both generators are connected in three generation cycles and can be optimized in an end-to-end fashion. 
Extensive qualitative and quantitative experimental results on four challenging generative tasks demonstrate that the proposed C2GAN is effective to generate photorealistic images with convincing details.

\section*{Acknowledgments}
This work was supported by the EU H2020 AI4Media No. 951911 project, by the Italy-China collaboration project TALENT:2018YFE0118400, and by the PRIN project PREVUE.



\small
\bibliographystyle{IEEEtran}
\bibliography{ref}


\begin{IEEEbiography}[{\includegraphics[width=1in,height=1.25in,clip,keepaspectratio]{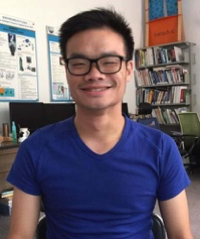}}]{Hao Tang}
is a Ph.D. candidate in the Department of Information Engineering and Computer Science at the University of Trento. He received the Master degree in computer application technology in 2016 at the School of Electronics and Computer Engineering, Peking University.
He was a visiting scholar in the Department of Engineering Science at the University of Oxford, from 2019 to 2020. His research interests are deep learning, machine learning and their applications to computer vision.
\end{IEEEbiography}


\begin{IEEEbiography}[{\includegraphics[width=1in,height=1.25in,clip,keepaspectratio]{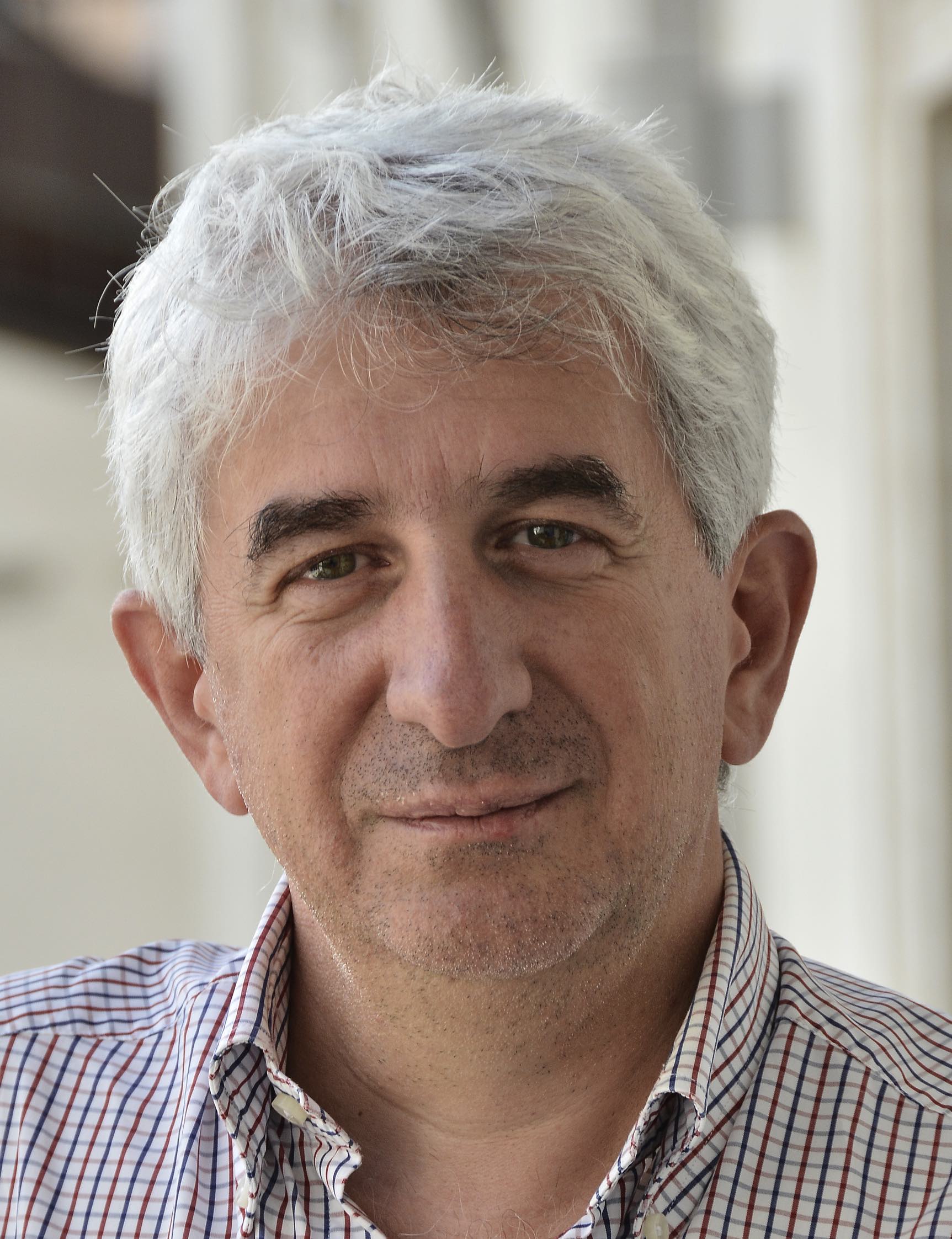}}]{Nicu Sebe} 
is Professor in the University of Trento, Italy, where he is leading the research in the areas of multimedia analysis and human behavior understanding. He was the General Co-Chair of the IEEE FG 2008 and ACM Multimedia 2013.  He was a program chair of ACM Multimedia 2011 and 2007, ECCV 2016, ICCV 2017 and ICPR 2020.  He is a general chair of ACM Multimedia 2022 and a program chair of ECCV 2024. He is a fellow of IAPR.
\end{IEEEbiography}

%
%
%
%
%




\end{document}